\documentclass[10pt,twocolumn,letterpaper]{article}

\usepackage[pagenumbers]{cvpr} 

\usepackage[dvipsnames]{xcolor}

\newcommand{\pxy}[1]{\textcolor{blue}{#1}}

\definecolor{cvprblue}{rgb}{0.21,0.49,0.74}
\usepackage[pagebackref,breaklinks,colorlinks,citecolor=cvprblue]{hyperref}
\usepackage{makecell}
\usepackage{multirow}

\def\paperID{2724} 
\def\confName{CVPR}
\def\confYear{2024}

\title{More Than Positive and Negative: Communicating Fine Granularity \\ in Medical Diagnosis}

\author{Xiangyu Peng\\
National University of Singapore \\
{\tt\small xiangyupeng@comp.nus.edu.sg}
\and
Kai Wang\\
National University of Singapore \\
{\tt\small kai.wang@comp.nus.edu.sg}
\and
Jianfei Yang\\
Nanyang Technological University \\
{\tt\small jianfei.yang@ntu.edu.sg}
\and
Yingying Zhu\\
University of Texas Arlington\\
{\tt\small yingying.zhu@uta.edu}
\and
Yang You\\
National University of Singapore\\
{\tt\small youy@comp.nus.edu.sg}
}

\begin{document}
\maketitle
\begin{abstract}
With the advance of deep learning, much progress has been made in building powerful artificial intelligence (AI) systems for automatic Chest X-ray (CXR) analysis. Most existing AI models are trained to be a binary classifier with the aim of distinguishing positive and negative cases. However, a large gap exists between the simple binary setting and complicated real-world medical scenarios. In this work, we reinvestigate the problem of automatic radiology diagnosis. We first observe that there is considerable diversity among cases within the positive class, which means simply classifying them as positive loses many important details. This motivates us to build AI models that can communicate fine-grained knowledge from medical images like human experts. To this end, we first propose a new benchmark on fine granularity learning from medical images. Specifically, we devise a division rule based on medical knowledge to divide positive cases into two subcategories, namely atypical positive and typical positive. Then, we propose a new metric termed AUC$^\text{FG}$ on the two subcategories for evaluation of the ability to separate them apart. With the proposed benchmark, we encourage the community to develop AI diagnosis systems that could better learn fine granularity from medical images. Last, we propose a simple risk modulation approach to this problem by only using coarse labels in training. Empirical results show that despite its simplicity, the proposed method achieves superior performance and thus serves as a strong baseline.
\end{abstract}

\section{Introduction}
\label{sec:intro}

Chest X-ray (CXR) is one of the most popular medical imaging technologies in the world for radiology diagnosis~\cite{esteva2019guide,litjens2017survey}. These imaging studies play a pivotal role in the assessment of various thoracic conditions, including lung infections, heart abnormalities, and skeletal disorders~\cite{franquet2001imaging, cherian2005standardized}. Usually, CXR is accompanied by a professional-made medical report of rich information, which comprehensively describes the findings on the patient to facilitate the following intervention or treatment.

\begin{figure}
    \centering
    \includegraphics[width=\linewidth, trim=7.6cm 2.5cm 9.6cm 1cm, clip]{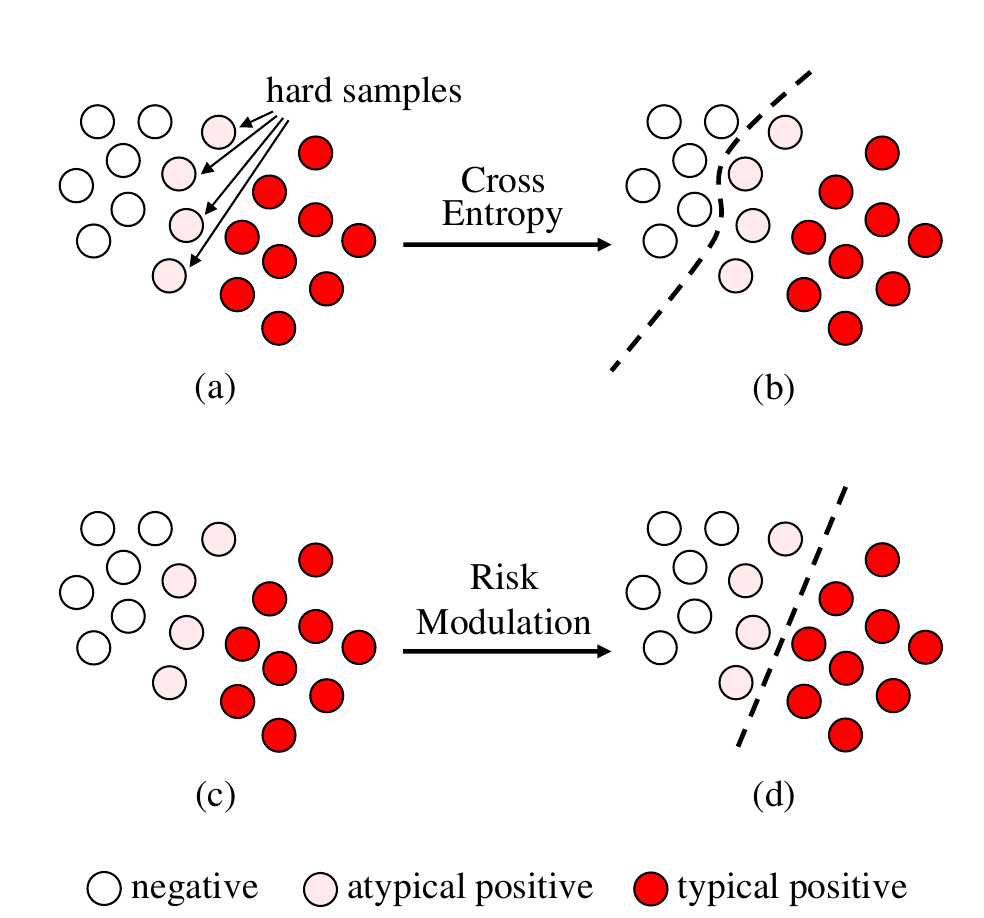}
    \caption{Comparison of the Cross-Entropy (CE) loss and our method in medical image diagnosis. The CE loss treats positive samples as hard samples, which leads to model overfitting. Moreover, it ignores the discrepancy between atypical positive and typical positive. We propose to modulate the risk in the CE loss for hard samples, which prevents the model from overfitting and improves its ability to learn fine granularity within the positive class.}
    \label{fig:hard}
\end{figure}


Considering the expertise, labor, and extensive time required in manual CXR reading, large-scale CXR datasets have been built to boost the development of artificial intelligence (AI) models for automatic analysis of CXRs~\cite{irvin2019chexpert, johnson2019mimic, johnson2019mimicjpg, wang2017chestx}. Following the longstanding natural image recognition~\cite{krizhevsky2009learning, deng2009imagenet}, these datasets extract from reports category labels for each radiograph, which are typically positive and negative. Such large-scale data with annotations largely supports researchers in developing AI diagnosis models by leveraging the advances in deep learning~\cite{lecun2015deep, he2016deep, krizhevsky2012imagenet}. So far, a number of powerful AI models have been proposed, some of which achieve or even surpass expert-level performance in radiology diagnosis~\cite{yuan2021large, pham2021interpreting}.

\begin{figure*}
    \centering
    \includegraphics[width=0.95\textwidth, trim=4.4cm 6cm 7.3cm 6cm, clip]{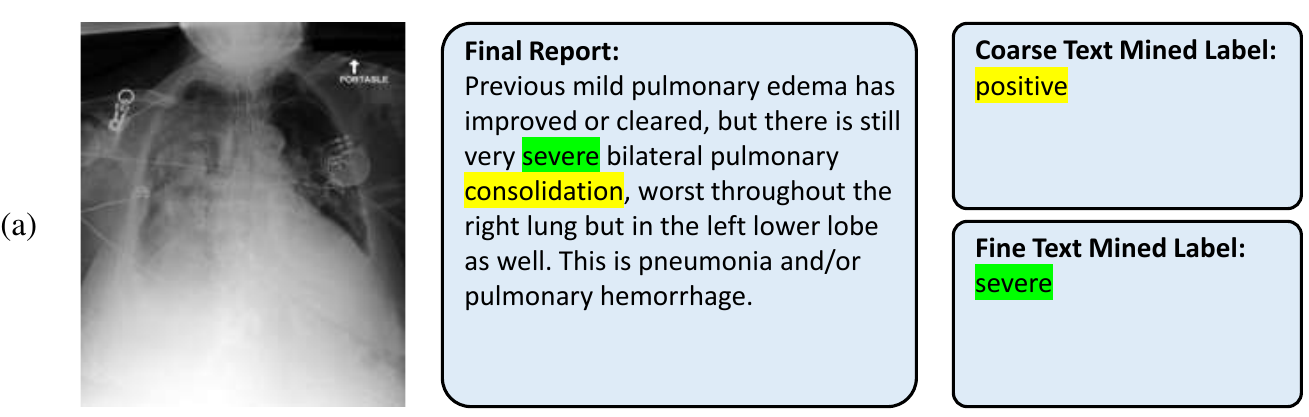}
    \includegraphics[width=0.95\textwidth, trim=4.4cm 6cm 7.3cm 6cm, clip]{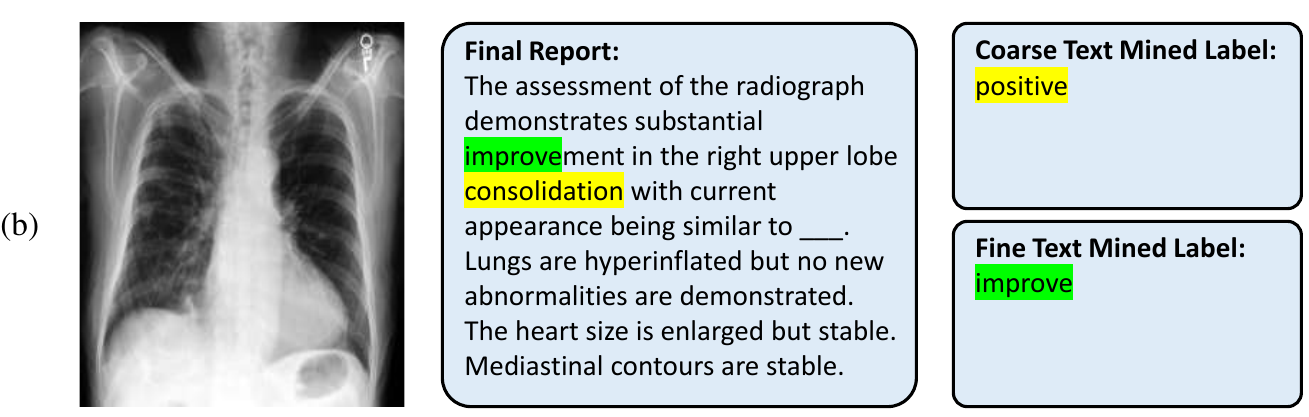}
    \includegraphics[width=0.95\textwidth, trim=4.4cm 6cm 7.3cm 6cm, clip]{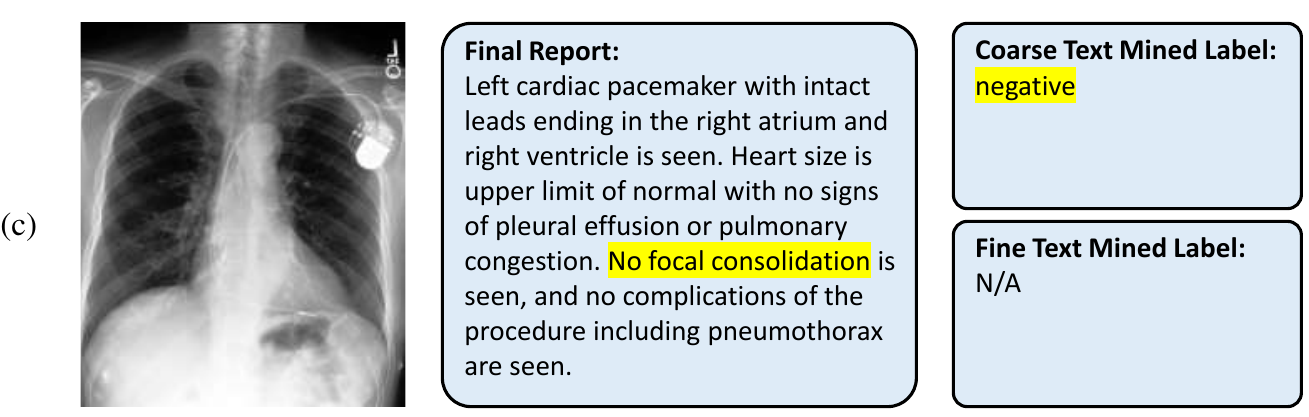}
    \caption{Example CXR images with correspondent reports and text-mined labels. Although subfigure (a) and subfigure (b) are both labeled as positive, there is a large discrepancy between them. Subfigure (a) is a typical positive case with severe symptoms of consolidation, while subfigure (b) is an atypical case in which the condition of the patient has substantially improved with little symptom left. Though labeled as positive, the case in (b) is closer to the fringe of negative in human expert evaluation.}
    \label{fig:moti}
\end{figure*}

Despite their remarkable performances, a problem with these models is the assumption that the CXR image for analysis is either positive or negative. This is true with some natural image classification tasks, where the image category is relatively clear~\cite{krizhevsky2009learning, deng2009imagenet}. However, unlike a coin is either heads or tails, the boundary between positive and negative outcomes in medical scenarios is not always evident.  For example, the report ``there is still very severe bilateral pulmonary consolidation'' of Fig.~\ref{fig:moti}(a) no doubt indicates a typical positive case for consolidation. However, Fig.~\ref{fig:moti}(b) with the description ``substantial
improvement in the right upper lobe consolidation'' shows that the condition of the patient has substantially improved with barely any symptoms left. Though also extracted as positive, the latter is closer to the fringe of negative in human expert evaluation.


Ignoring such a discrepancy within the positive class could lead to the misalignment between the perception of the model and the evaluation of human experts.
As shown in Fig.~\ref{fig:hard}(a), traditional methods tend to use the Cross-Entropy (CE) loss to train a binary classifier~\cite{irvin2019chexpert}. Although training with the CE loss could classify typical negative and positive cases well, predictions on atypical positive cases are likely to be biased. This is because some atypical positive samples with high losses are treated as hard samples. By fitting these hard samples, the model tends to learn an overfitting decision boundary (Fig.~\ref{fig:hard}(b)), resulting in inferior predictions in real-world practice. 

In this work, we aim to build an AI model that could communicate fine-grained information from medical images for better diagnostic practices. To this end, we propose a new benchmark for fine granularity learning in medical images. Specifically, we first design a medically plausible division rule to further divide positive cases into two subcategories, namely atypical positive and typical positive, according to the report description.  Then, we propose a new metric named AUC$^\text{FG}$ for the  evaluation of fine granularity learning. AUC$^\text{FG}$ calculates the AUC between atypical and typical positive cases, with atypical positive being 0 and typical positive being 1. The proposed benchmark would help future AI diagnosis systems to learn discrepancy among positive cases.

Last, we propose a simple yet effective approach to fine granularity learning with only coarse labels. We propose to modulate the risk in the CE loss function, which results in the Partially Huberised Cross-Entropy (PCE) loss~\cite{feng2021can}. The PCE loss has a lower risk for hard samples in the low-confidence region. By reducing the risk for hard samples, we can prevent the model from overfitting and expect a decision boundary that is more aligned with human expert evaluation (Fig.~\ref{fig:hard}(c) \& Fig.~\ref{fig:hard}(d)). With such a decision boundary, the model could better distinguish the discrepancy among complicated clinical cases. We conduct experiments on the large-scale CXR dataset MIMIC-CXR-JPG~\cite{johnson2019mimicjpg} to validate the proposed method. We choose two pathologies of clinical significance and prevalence for validation, which are consolidation and edema. Experimental results show that, despite its simplicity, our method achieves superior performance under the metric AUC$^\text{FG}$ and thus could serve as a strong baseline.


Our main contributions can be summarized as:
\begin{itemize}
    \item We identify the deficiency of learning a simple binary classifier for medical image analysis and propose that fine-grained knowledge should be taken into account when building an AI diagnosis system.
    \item We devise a division rule to divide the positive class into atypical positive and typical positive based on the report. With this division rule, we propose a new metric AUC$^\text{FG}$ for evaluation of fine granularity learning in medical image analysis.
    \item We propose a simple yet effective risk modulation approach that allows a model to learn fine granularity from coarsely annotated samples.
    \item We conduct empirical studies of the proposed method on the large-scale MIMIC-CXR-JPG and demonstrate its superior performance.
\end{itemize}

\section{Related Works}
\label{sec:related}

\subsection{Medical Image Analysis}
Recent years have seen much effort in the creation of large-scale chest X-ray datasets, which is instrumental in propelling the progress of AI diagnosis models within the realm of medical imaging. A pioneer work is done by Wang \etal~\cite{wang2017chestx}, who proposed the ChestX-ray14 dataset comprising 112,120 CXR images with multi-class labels for 14 common thorax diseases. Later, CXR datasets with larger scales are proposed, such as CheXpert~\cite{irvin2019chexpert} which contains more than 200,000 scans, and MIMIC-CXR~\cite{johnson2019mimic} which comprises more than 350,000 images.

The sheer volume and variety of data in these datasets facilitate the research on developing robust and generalizable AI diagnosis models. So far, a large number of sophisticated AI models have been proposed, some of which achieve or even surpass expert-level performance on many medical image interpretation tasks. Rajpurkar \etal conducted an early work on the application of deep learning techniques to enhance pneumonia detection in chest X-rays~\cite{rajpurkar2017chexnet}. The authors proposed CheXNet, a convolutional neural network (CNN) that was trained on the ChestX-ray14 dataset and achieved performance levels comparable to human radiologists. Rajpurkar \etal further developed an improved model of CheXNet called CheXNeXt~\cite{rajpurkar2018deep}. By incorporating a two-stage training and ensemble, CheXNeXt achieved radiologist-level performance on 11 pathologies out of 14 on the ChestX-ray14 dataset. Recently, Irvin \etal introduced various approaches to handling uncertain labels on the CheXpert dataset to cater to the characteristics of different pathologies~\cite{irvin2019chexpert}. Pham \etal exploited hierarchical disease dependencies~\cite{chen2019deep, yao2017learning} and the label smoothing technique~\cite{muller2019does} to train CNN models which achieved state-of-the-art performance on CheXpert. All of these methods, however, ignore the discrepancy among the same class of medical images. In this work, we propose to make AI models learn the fine granularity of medical images and thus reduce the gap between the perception of an AI model and human expert evaluation.

\subsection{Fine Granularity Leaning}
Traditional natural image classification is usually carried out on categories with relatively definite concepts such as ``cats'' and ``cars''~\cite{krizhevsky2009learning, deng2009imagenet}. In the medical scenario, however, the boundary between positive and negative is not so obvious. There could even be a large discrepancy within the same positive class. Hence, learning fine granularity from medical images is of practical significance for an AI diagnosis model.

In the literature, various approaches have been proposed to extract detailed information from medical images. Guan~\etal proposed AG-CNN, an attention-guided CNN for thorax disease classification~\cite{guan2018diagnose}. The attention mechanisms used in this work enable the model to focus on relevant regions in the input images and learn more discriminative features for accurate disease classification. Tang~\etal introduced an attention-guided curriculum learning framework for weakly supervised classification and localization of thoracic diseases~\cite{tang2018attention}, where confident samples with their class-conditional heatmaps are iteratively extracted to guide the learning of more distinctive features. Ye~\etal proposed Probabilistic Class Activation Map (PCAM) pooling to address the problem of lesion localization with only image-level supervision~\cite{ye2020weakly}. In this work, we propose risk modulation as a simple yet effective approach to learning fine granularity without using attention mechanisms or fine-grained annotations.

\section{A New Benchmark} \label{sec:benchmark}
Most previous works simply view CXR diagnosis as a binary classification problem, which ignores the complexity in real-world practice. In this section, we introduce a new benchmark for learning fine-grained knowledge from medical images. 

\subsection{Data Preparation}
We build the benchmark on the large-scale CXR dataset MIMIC-CXR-JPG~\cite{johnson2019mimicjpg}, which consists of CXR images paired with medical reports. Traditionally, category labels for each finding are extracted from reports and the possible categories are positive ($1$), negative ($0$), uncertain ($u$), and blank ($blank$) if a finding is not mentioned in the report. However, such coarse category labels lose much information in the original report. Therefore, we use fine-grained labels provided by Zhang \etal~\cite{zhang2023expert}. The authors extract from reports multiple levels of disease severity, as well as traditional categories of negative, uncertain, and positive. In addition to that, we further extract keywords on change in time such as ``improve'' and ``worsen'' from reports, using the same extraction tool as in~\cite{zhang2023expert}. The multiple severity levels and keywords on change in time jointly compose the fine-grained labels for the proposed benchmark. We then build a new metric upon the fine-grained labels for evaluation of fine granularity learning, which is elaborated in Sec.~\ref{sec:metric}.

\begin{table*}[t]
    \centering
    \begin{tabular}{c|c|c}
        \Xhline{1.5pt}
        \multicolumn{3}{c}{\textbf{(a) Division based on severity}} \\
        \hline
        & \makecell{Fine \\ Text-Mined Label} & Example Report \\
        \hline
        \multirow{6}{*}[-1.6em]{\makecell{atypical \\ positive}} & trace & \makecell[l]{There is a new \textbf{trace} pleural effusion on the left.} \\
        \cline{2-3}
        & small & \makecell[l]{There is a \textbf{small} right pleural effusion and atelectasis at the right lung base.} \\
        \cline{2-3}
        & tiny & \makecell[l]{There is scattered atelectasis and there are \textbf{tiny} bilateral pleural effusions.} \\
        \cline{2-3}
        & minor & \makecell[l]{The right hemithorax segmental atelectasis adjacent to the \textbf{minor} fissure \\ is stable.} \\
        \cline{2-3}
        & minimal & \makecell[l]{Lung volumes are improved with \textbf{minimal} bibasilar atelectasis.} \\
        \cline{2-3}
        & mild & \makecell[l]{There is \textbf{mild} pulmonary vascular congestion and interstitial edema.} \\
        \cline{2-3}
        & mild-to-moderate & \makecell[l]{In comparison with the study of \underline{\hspace{0.5cm}}, there is continued substantial \\ cardiomegaly with \textbf{mild-to-moderate} pulmonary edema.} \\
        
        \hline 
        \multirow{4}{*}[-4.2em]{\makecell{typical \\ positive}} & large & \makecell[l]{There is a \textbf{large} heterogeneous consolidation of the right lower lung with \\ air bronchograms compatible with pneumonia.} \\
        \cline{2-3}
        & massive & \makecell[l]{Frontal and lateral radiographs of the chest demonstrate persistent \textbf{massive} \\ left-sided pleural effusion, occupying at least two-thirds of the left hemithorax.} \\
        \cline{2-3}
        & substantial & \makecell[l]{Diffuse bilateral pulmonary opacifications are again seen, consistent with \\ \textbf{substantial} pulmonary edema.} \\
        \cline{2-3}
        & moderate & \makecell[l]{There continues to be a moderate left effusion.} \\
        \cline{2-3}
        & moderate-to-severe & \makecell[l]{There is interval development of \textbf{moderate-to-severe} interstitial pulmonary \\ edema with some element of alveolar edema and bilateral pleural effusions,} \\
        \cline{2-3}
        & acute & \makecell[l]{\textbf{Acute} pulmonary congestion with central pulmonary edema and left-sided \\ pleural effusion.} \\
        \cline{2-3} 
        & severe & \makecell[l]{Very \textbf{severe} pulmonary consolidation is seen in the right lung.} \\

        \Xhline{1.5pt}
        \multicolumn{3}{c}{\textbf{(b) Division based on change in time}} \\
        \hline
        & \makecell{Fine \\ Text-Mined Label} & Example Report \\
        
        \hline
        \multirow{4}{*}[-1.8em]{\makecell{atypical \\ positive}} & clear & \makecell[l]{Since \underline{\hspace{0.5cm}}, the volume of consolidation in the right mid and \\ lower lung zone has decreased and pulmonary edema has largely \textbf{clear}ed.} \\
        \cline{2-3}
        & resolution & \makecell[l]{Cardiomegaly is accompanied by improved pulmonary vascular congestion \\ and near \textbf{resolution} of interstitial edema. } \\
        \cline{2-3}
        & improve & \makecell[l]{The left hemidiaphragm and left costophrenic sulcus are now better defined, \\ 
        suggesting interval \textbf{improvement} in the left lower lobe collapse/consolidation.} \\
        \cline{2-3}
        & decrease & \makecell[l]{As compared to the previous radiograph, the signs of interstitial lung edema \\ have \textbf{decrease}d in extent and severity.} \\

        \hline
        \multirow{3}{*}[-0.7em]{\makecell{typical \\ positive}} & progress & \makecell[l]{There is interval \textbf{progress}ion of vascular congestion and interstitial \\ pulmonary edema.} \\
        \cline{2-3}
        & worsen & \makecell[l]{Moderate to severe pulmonary edema has \textbf{worsen}ed.} \\
        \cline{2-3} 
        & increase & \makecell[l]{Since the chest radiographs obtained 3 days prior, there has been \\ a significant \textbf{increase} in left lung atelectasis with leftward mediastinal shift.} \\
        
        \Xhline{1.5pt}
    \end{tabular}
    \caption{Details of the proposed division rule for the two subcategories of atypical and typical positive. We first extract fine-grained labels from reports and then make the division based on two dimensions: (a) division based on severity; (b) division based on change in time. Generally, a slight symptom is classified as atypical positive (\eg, ``mild'') and a significant one is considered typical positive (\eg, ``severe''). Moreover, change to a better condition is considered atypical positive (\eg, ``improve''), while change to a worse condition is viewed as typical positive (\eg, ``worsen'').}
    \label{tab:atypical}
\end{table*}

\subsection{Evaluation Metric} \label{sec:metric}
We first carefully devise a division rule to divide the positive class into two subcategories, namely atypical positive and typical positive. The rationality of the division rule has been validated by human experts. We show the details of the division rule are in Tab.~\ref{tab:atypical}. Specifically, we make the division based on two dimensions, which are severity and change in time. On the one hand, for severity, slight symptoms are classified as atypical positive (\eg, ``small", ``minor'', ``mild''), while significant symptoms are considered typical positive (\eg, ``severe'', ``acute''). On the other hand, change to a better condition is considered atypical positive (\eg, ``improve'', ``decrease''), while change to a worse condition is considered typical positive (\eg, ``worsen'', ``increase''). If a sample has no fine-grained labels, it is treated as typical positive (\eg, ``There is consolidation in the right lung.'').

Based on the division rule, we then propose a new metric termed AUC$^\text{FG}$. AUC$^\text{FG}$ calculates the AUC between the two subcategories, with atypical positive being 0 and typical positive being 1. With the proposed metric, we are able to evaluate how well a model can discover the fine-grained knowledge in medical images. A higher AUC$^\text{FG}$ indicates a better ability to distinguish atypical and typical positive and a better alignment with human expert evaluation.
\section{Method}
\label{sec:method}

In this section, we introduce a simple yet effective approach named PU-RM for fine granularity learning. When training this model,  we \textbf{do not} need fine-grained labels.

\subsection{Problem Formulation}
We consider a dataset of chest radiographs $X=\{x_i\}^N_{i=1}$ and correspondent labels $Y=\{y_i\}^N_{i=1}$ (\eg, MIMIC-CXR-JPG), where $y_i = (y^1_i, y^2_i, ..., y^C_i)$ is a $C$ dimensional vector that contains labels for $C$ findings such as symptoms and diseases. Here, each label $y^c$ could be one of positive ($1$), negative ($0$), uncertain ($u$), and blank ($blank$) if finding $c$ is not mentioned in the report. 

Given medical images with coarse category labels, we aim to train a model $f_{\theta}$ with parameters $\theta$ capable of communicating fine-grained knowledge in the same class. Although the labels are in the form of multi-label classification~\cite{durand2019learning}, we apply our method to each finding individually. This rules out the potential influence of multi-label learning and thus better focuses on the effect of the method itself. Therefore, the output of the model for an observation is $f_{\theta}(x) = (p^n, p^p)$, where $p^n$/$p^p$ represents the probability of the sample being negative/positive. Note that we apply softmax before outputting the probabilities to make sure that $p^n + p^p = 1$. 

\subsection{Fine Granularity Learning}
Most previous works simplify radiology diagnosis as a binary classification problem. A common approach is to use the Cross-Entropy (CE) loss for training, which is formulated as 
\begin{equation}
    \mathcal{L}_{CE}(y^c) = - (1 - y^c) \log (p^n) - y^c \log (p^p)
\end{equation}
Here, $y^c$ is either $0$ (negative) or $1$ (positive), and we elaborate on how to deal with $y^c = u$ in Sec.~\ref{sec:method_uc}. Although the CE loss can separate easy positive and negative samples, it may not classify borderline cases well due to heavy penalization on hard samples. In radiology diagnosis, however, such a heavy penalization on hard positive samples is not expected, as these hard samples include many atypical positive cases which are arguably closer to negative. Training with the CE loss may lead to model overfitting and misalignment between model perception and human expert evaluation. 


\begin{figure}
    \centering
    \includegraphics[width=\linewidth, trim=0.8cm 0cm 0.7cm 0cm, clip]{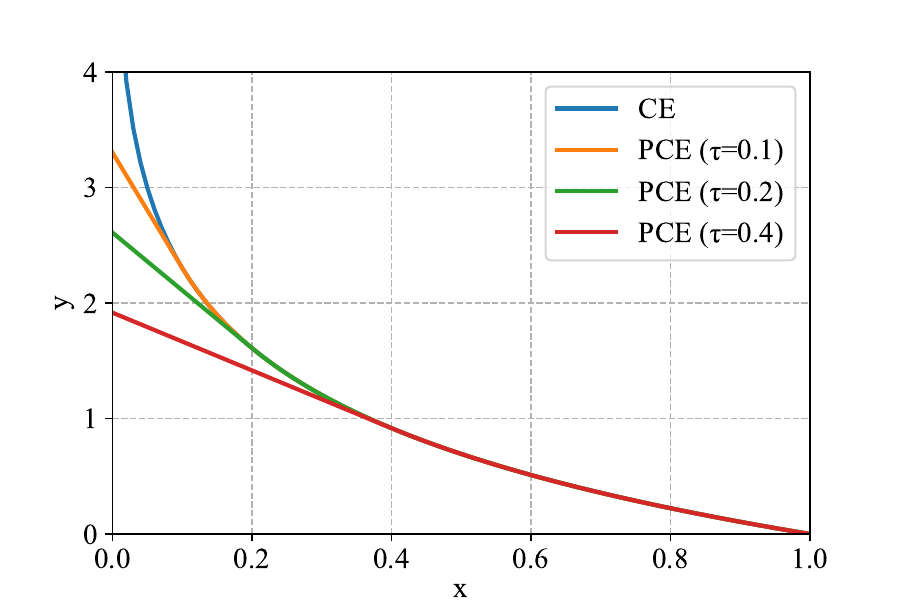}
    \caption{Loss curves for the CE loss and PCE losses with different parameters $\tau$.}
    \label{fig:CE}
\end{figure}

To address this issue, we propose to modulate the risk in the loss function for hard samples. Specifically, we use the Partially Huberised Cross Entropy (PCE) loss~\cite{feng2021can} as the training target. The PCE loss can be formulated as 
\begin{multline} \label{eq:pce}
    \mathcal{L}_{PCE} (y^c=1) = \\
    \begin{cases}
        - \frac{1}{\tau} (p^p - \tau) - \log \tau, & \text{if} \: p^p \leq \tau \\
        -\log p^p, & \text {otherwise}
    \end{cases},
\end{multline}
where $\tau \in (0, 1)$ is the tangent point of the logarithmic curve. As shown in Fig.~\ref{fig:CE}, the PCE loss has a smaller risk and gradient than the CE loss in the low-confidence region (\ie, $\text{x} < \tau$). In this way, we can prevent the model from overfitting on hard samples, thus distinguishing atypical and typical cases better. We refer to applying risk modulation on positive samples as P-RM.

The PCE loss was originally proposed to deal with the label noise problem~\cite{feng2021can}. It is demonstrated to be noise-robust due to its property of bounded risk in the presence of noisy labels~\cite{feng2021can}. In this work, we use the same loss function but for a different purpose \pxy{}. We use this risk-bounded loss to show that risk modulation facilitates fine granularity learning within a single class.

\subsection{Uncertainty Approaches}
\label{sec:method_uc}
Due to the inherent ambiguities in medical reports, labels extracted from them could be uncertain (\ie, $y^c = u$) in addition to positive and negative. How to deal with these uncertainty labels varies from work to work and still remains an open problem~\cite{irvin2019chexpert, pham2021interpreting, yuan2021large}. The most common approaches to using uncertainty labels include:
\begin{itemize}
    \item \textbf{Ignoring} (U-Ignore). This approach simply ignores uncertain labels. In multi-label classification, this is achieved by masking losses with respect to samples with uncertainty labels. In this work, as we conduct training for each observation individually, we simply remove samples with uncertainty labels in training for each observation.
    \item \textbf{Binary Mapping}. In this approach, uncertainty labels are all mapped to positive (U-Ones) or all mapped to negative (U-Zeros) as a whole. Compared to U-ignore, this approach takes advantage of the uncertain instances. However, label noises are inevitably introduced in the mapping process, which may mislead the model training and degrade the performance instead~\cite{irvin2019chexpert}.
\end{itemize}

In this work, we propose a new approach termed U-RM to better handle uncertain labels. We map all uncertain labels to positive and then apply the risk-modulated loss (\ie, the PCE loss) to them. Similar to fine granularity learning with positive samples, this approach also reduces the risk for uncertain cases that are more likely to be negative. As a result, the model is trained to learn fine granularity from uncertain cases.

\subsection{Training Objective}
The overall training objective incorporates losses on negative, uncertain, and positive samples, which is then written as
\begin{equation}
\begin{split}
    \mathcal{L} = 
     & 1_{[y^c=0]} \cdot \mathcal{L}_{CE}(0) + \\
     & 1_{[y^c=u]} \cdot \mathcal{L}_{PCE}(1) + \\
     & 1_{[y^c=1]} \cdot \mathcal{L}_{PCE}(1)
\end{split}
\end{equation}
Since we apply risk modulation to both positive and uncertain samples, we refer to our method as PU-RM. We set the hyper-parameter $\tau=0.3$ for the PCE loss. Ablation studies on $\tau$ are presented in Sec.~\ref{sec:abl}.

\section{Experiments}
\label{sec:exp}

\subsection{Data}
We evaluate our method on the proposed benchmark described in Sec.~\ref{sec:benchmark}. We use the large-scale CXR dataset MIMIC-CXR-JPG~\cite{johnson2019mimicjpg}, which consists of JPG format files derived from the original DICOM images in MIMIC-CXR~\cite{johnson2019mimic}. Officially, the MIMIC-CXR-JPG dataset is organized into a training set of 222,758 studies from 64,586 patients, a validation set of 1808 studies from 500 patients, and a test set of 3269 studies from 293 patients. Each study consists of a medical report and one or more Chest X-rays. In this work, we only consider CXR images with anteroposterior (AP) or posteroanterior (PA) views. Samples with blank labels are also ignored. We use the official split of MIMIC-CXR-JPG for training and validation.


\subsection{Experimental Setup}
\paragraph{Training.}
For all experiments in this work, we use the same backbone network and training process for fair comparison. We choose DenseNet121~\cite{huang2017densely} as the backbone network, following the CheXpert paper~\cite{irvin2019chexpert}. The backbone network is initialized with ImageNet pre-trained weights\footnote{We use the official pre-trained weights provided by PyTorch. See \url{https://pytorch.org/vision/0.10/_modules/torchvision/models/densenet.html}.} for faster convergence except for the last fully connected layer which is randomly initialized. The input images are rescaled to $224 \times 224$ pixels. We use the Adam~\cite{kingma2014adam} optimizer for training with $\beta$-parameters of $\beta_1 = 0.9$, $\beta_2 = 0.999$, a fixed learning rate of $1e^{-4}$ and a weight decay of $0$. The model is trained for 50,000 iterations with a total batch size of 32 on 2 GPUs. In this way, all experiments have the same amount of training. We save checkpoints every 10,000 iterations, obtaining 5 checkpoints in each run.

\begin{table}[t]
    \centering
    \begin{tabular}{c|c c}
        \toprule
        & Consolidation &  Edema\\
        \midrule
        U-Ignore  & 0.7481 $\pm$ 0.0095 & 0.6369 $\pm$ 0.0089 \\
        U-Zeros  & 0.7675 $\pm$ 0.0058 & 0.5565 $\pm$ 0.0013 \\
        U-Ones  & 0.7716 $\pm$ 0.0144 & 0.6374 $\pm$ 0.0123 \\
        PU-RM (Ours) & \textbf{0.8038} $\pm$ 0.0078 & \textbf{0.6663} $\pm$ 0.0094 \\
        \bottomrule
    \end{tabular}
    \caption{AUC$^\text{FG}$ results on the validation set. We report mean $\pm$ std of three runs. Best results are marked in bold.}
    \label{tab:main_res}
\end{table}

\paragraph{Validation.}
For each method, we run the model three times, each with a different seed. We choose the best checkpoint on the validation set from each run and report the mean and standard deviation of the three best checkpoints. We validate our method on consolidation and edema individually, two pathologies of clinical importance and prevalence. Note that only samples with positive labels are involved in validation since our goal is to learn fine-grained knowledge within the positive class.

\begin{table}[]
    \centering
    \begin{tabular}{c|c|c}
        \toprule
        & Consolidation & Edema \\
        \midrule
        U-Ignore & 0.7481 $\pm$ 0.0095 & 0.6369 $\pm$ 0.0089 \\
        U-Ignore + P-RM & 0.7728 $\pm$ 0.0127 & 0.6425 $\pm$ 0.0169 \\
        PU-RM (Ours) & \textbf{0.8038} $\pm$ 0.0078 & \textbf{0.6663} $\pm$ 0.0094 \\ 
        \bottomrule
    \end{tabular}
    \caption{Results for applying risk modulation (RM) to the uncertain and the positive class individually. Best results are marked in bold.}
    \label{tab:abl_pu}
\end{table}

\begin{figure}[t]
    \centering
    \begin{subfigure}[b]{\linewidth}
        \centering
        \includegraphics[width=\linewidth]{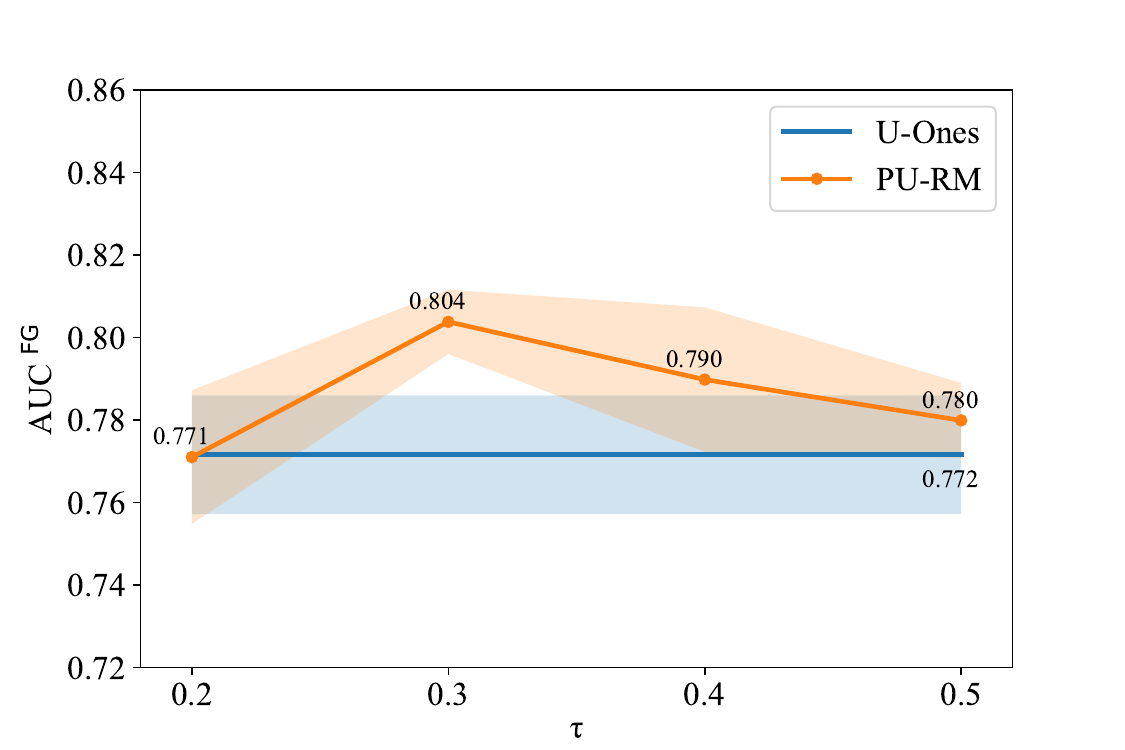}
        \caption{Results of different $\tau$ for consolidation.}
        \label{fig:tau_con}
    \end{subfigure}
    \hfill
    \begin{subfigure}[b]{\linewidth}
        \centering
        \includegraphics[width=\linewidth]{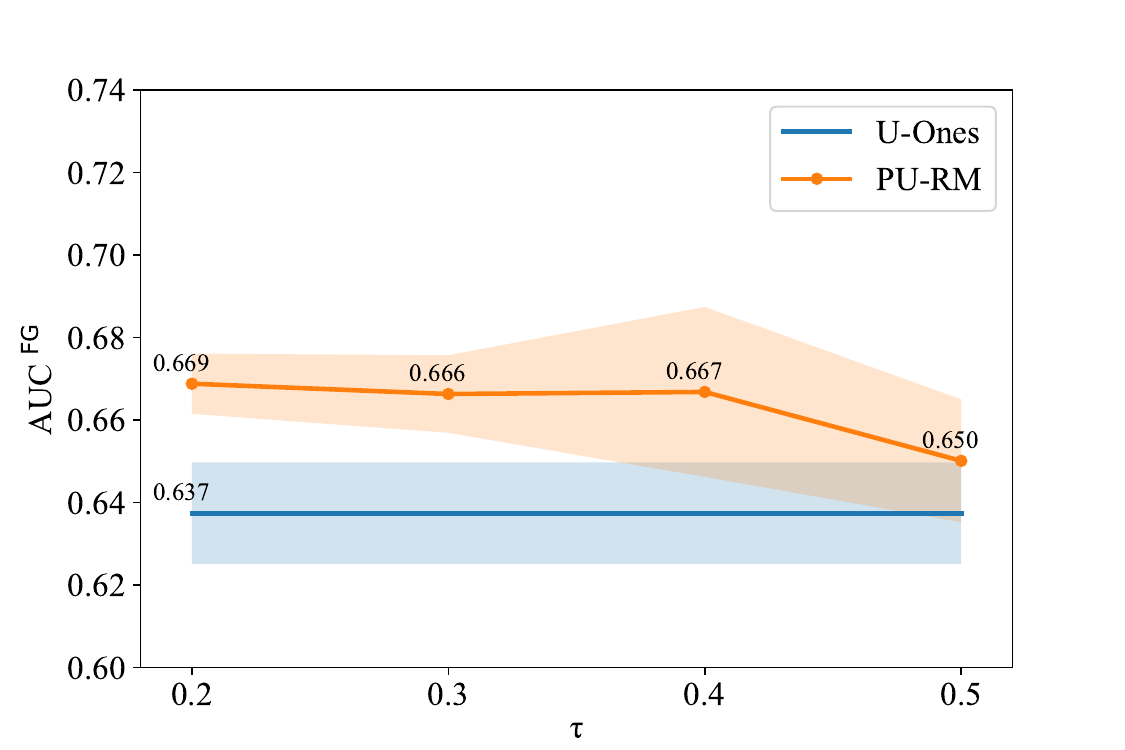}
        \caption{Results of different $\tau$ for edema.}
        \label{fig:tau_ede}
    \end{subfigure}
    \caption{Ablation studies of different $\tau$ on the validation set for consolidation and edema. The line and the band represent the mean and std of three runs respectively.}
    \label{fig:tau}
\end{figure}

\begin{figure*}[t]
    \centering
    \includegraphics[width=0.95\textwidth, trim=3.5cm 5cm 3.2cm 5cm, clip]{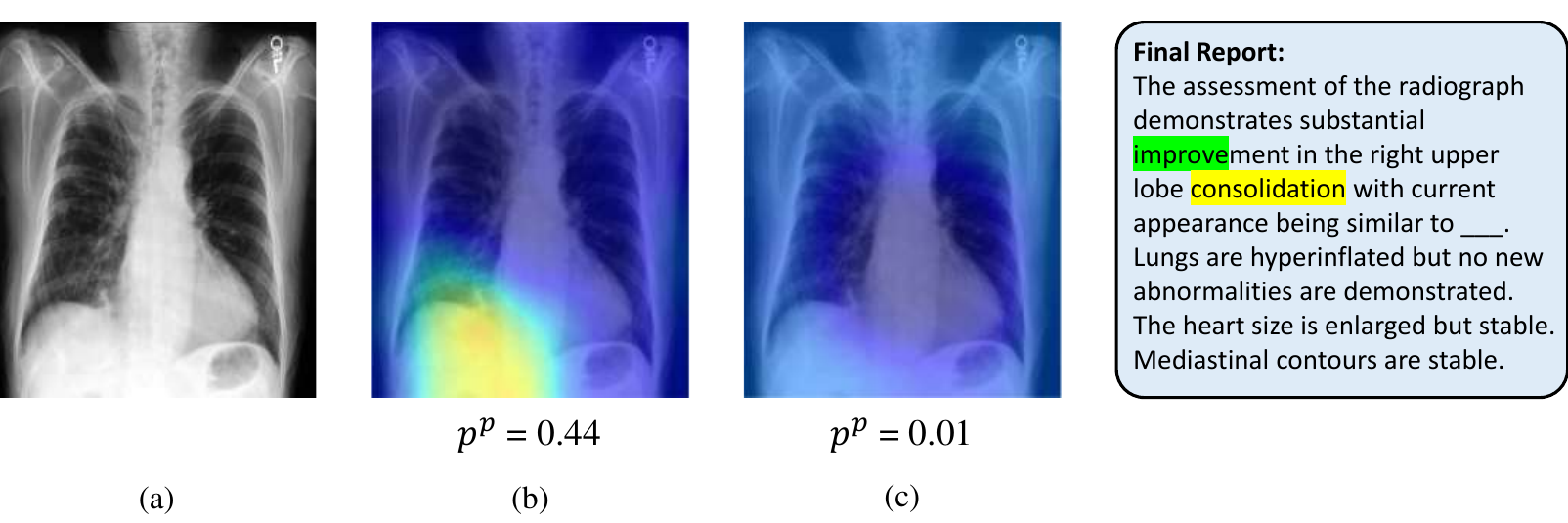}
    \caption{Visualization of model predictions using CAM. Subfigure (a) shows the original medical image, which is considered atypical positive according to the proposed division rule in Tab.~\ref{tab:atypical}. The baseline U-Ones model trained with the CE loss falsely activates in the right lower lobe area as shown in subfigure (b). In contrast, our method shown in subfigure (c) does not focus on any specific area. It also predicts a much lower positive probability of 0.01 than the baseline model of 0.44.}
    \label{fig:cam}
\end{figure*}

\subsection{Main Results}
We compare our method to several baseline uncertainty approaches under the proposed metric AUC$^{\text{FG}}$ in Tab.~\ref{tab:main_res}. The baseline uncertainty approaches are trained with the CE loss by default if not otherwise specified. We report mean $\pm$ std of three best checkpoints on the validation set for each method. One can see that U-Ones achieves the best performance among the three baseline approaches. For edema, U-Ignore is almost as good as U-Ones but U-Zeros leads to a significant drop in performance. This is probably because most uncertain phrases for edema actually convey findings that are more positive such as ``likely superimposed pulmonary edema.'' However, such a contrast is not observed for consolidation. Both U-Zeros and U-Ones achieve higher performance than U-Ignore for consolidation, which indicates the information in uncertain consolidation samples is beneficial to learning fine-grained knowledge.


Finally, our method outperforms all the baseline approaches by a large margin.
It achieves AUC$^{\text{FG}}$ of 0.8038 for consolidation, which is more than 0.03 higher than U-Ones, and 0.6663 for Edema, which is more than 0.02 higher than U-Ones.
This demonstrates the effectiveness of risk modulation for learning fine granularity without fine-grained labels. We conduct ablation studies in Sec.\ref{sec:abl} to study the effect of applying risk modulation on positive and uncertain cases individually.



\subsection{Ablation Studies} \label{sec:abl}
\paragraph{Hyper-parameter $\tau$.} We show in Fig.~\ref{fig:tau} the results of varing $\tau$ in Eq.~\ref{eq:pce}. The line and the band represent the mean and the standard deviation respectively. For both consolidation and edema, our method outperforms the baseline method in a wide range of $\tau$ (from 0.3 to 0.5 for consolidation, from 0.2 to 0.5 for edema), while the std of our method is lower or on par with the baseline. This shows the robustness of our method towards the hyper-parameter.

\paragraph{Effect on Individual Classes.} We study the effect of applying the risk modulation loss on each individual class, \ie, the uncertain and the positive class. The results are shown in Tab.~\ref{tab:abl_pu}. One can see that for both consolidation and edema, applying the risk modulation loss only on positive samples achieves better results than the baseline when uncertain samples are ignored. This shows the effectiveness of risk modulation for learning fine-grained knowledge from positive samples. When the risk modulation is applied to uncertain samples in addition to positive samples, the performances are further improved, which demonstrates that U-RM can serve as an effective uncertainty approach.

\subsection{Visualization}
We visualize the prediction of models in Fig.~\ref{fig:cam} using Class Activation Mapping (CAM)~\cite{zhou2016learning}. Specifically, we retrieve the last activation map of convolution layers in the DenseNet-121 model, upscale it to the size of the original image, and overlay it on the image to get the final visualization result. As shown in Fig.~\ref{fig:cam}(b), the baseline U-Ones model trained with the CE loss produces a false activation in the right lower lobe area as a consequence of overfitting to hard samples. In contrast, our method in Fig.~\ref{fig:cam}(c) has a low and uniform activation in most areas and does not focus on any specific part. Moreover, the predicted positive probability of our method for such an atypical positive case is 0.01, which is much lower than the baseline approach of 0.44. Such a low score is better aligned with the human expert evaluation that the patient's condition has substantially improved with little symptom left.

\section{Conclusion}

In this work, we reinvestigate the problem of automatic medical image analysis. We identify that there is a large gap between simple binary outcomes from traditional AI diagnosis models and complicated real-world medical scenarios. To address the issue, we propose a new benchmark for learning fine granularity from medical images which could help future AI diagnosis systems to learn discrepancy among positive cases. Moreover, we propose a simple yet effective risk modulation approach to this problem. Experimental results show that the proposed method, despite its simplicity, achieves superior performance on the large-scale MIMIC-CXR-JPG dataset and thus serves as a strong baseline on the proposed benchmark.

{
    \small
    \bibliographystyle{ieeenat_fullname}
    \bibliography{main}
}


\end{document}